\documentclass[lettersize,journal]{IEEEtran}
\usepackage{amsmath,amsfonts}
\usepackage{algorithmic}
\usepackage{algorithm}
\usepackage{array}
\usepackage{textcomp}
\usepackage{stfloats}
\usepackage{url}
\usepackage{verbatim}
\usepackage{graphicx}
\usepackage{cite}

\usepackage{multirow} %
\usepackage{color,xcolor}
\usepackage{subcaption}
\usepackage{booktabs} 
\usepackage{graphbox} 
\usepackage{gnuplottex}
\usepackage{bm}
\usepackage{soul} 

\begin{document}

\title{Shape-Aware Monocular 3D Object Detection}

\author{Wei Chen, Jie Zhao, Wan-Lei Zhao*, Song-Yuan Wu
\thanks{Wei Chen and Wan-Lei Zhao are with the Department of Computer Science and Technology, Xiamen University, Xiamen 361005, China (e-mail: chenwei128@stu.xmu.edu.cn; wlzhao@xmu.edu.cn).}
\thanks{Jie Zhao and Song-Yuan Wu are with the Ningbo Boden AI Technology Co., Ltd., Ningbo 315048, China (e-mail: jie.zhao@bodenai.com; songyuan.wu@bodenai.com).} 
\thanks{This work has been submitted to the IEEE for possible publication. Copyright may be transferred without notice, after which this version may no longer be accessible.}
}

\markboth{}%
{Shell \MakeLowercase{\textit{et al.}}: A Sample Article Using IEEEtran.cls for IEEE Journals}

\maketitle

\begin{abstract}
The detection of 3D objects through a single perspective camera is a challenging issue. The anchor-free and keypoint-based models receive increasing attention recently due to their effectiveness and simplicity. However, most of these methods are vulnerable to the occlusion and truncation of objects. In this paper, a single-stage monocular 3D object detection model is proposed. An instance-segmentation head is integrated into the model training, which allows the model to be aware of the visible shape of a target object. Therefore, the detection largely avoids interference from irrelevant regions surrounding the target objects. In addition, we also reveal that the popular IoU-based evaluation metrics, which were originally designed for evaluating stereo or LiDAR-based detection methods, are insensitive to the improvement achieved by the monocular 3D object detection algorithms. A novel evaluation metric, namely average depth similarity (ADS) is proposed for the monocular 3D object detection models. Our method outperforms the comparison baseline in terms of both the popular and the proposed evaluation metrics while maintaining real-time efficiency. 
\end{abstract}


\section{Introduction}
The three-dimensional visual object detector is a core component of an autonomous driving system. The quality of the detection results has a direct impact on the subsequent tasks, such as tracking and driving planning. Recently, we have witnessed the great advances in the design of 3D object detection models. Most of them \cite{yang2018pixor,zhou2018voxelnet,yan2018second} rely on 3D LiDAR laser scanners. Although LiDAR point clouds allow the detectors to achieve accurate 3D scene localization, the hardware is too expensive to be equipped with normal cars. Stereo camera rigs are the alternative option in many 3D object detection models 
\cite{peiliang2019stereorcnn,sun2020disprcnn,wang2019pseudo}, where depth information is estimated by building accurate pixel correspondences between the left and the right cameras. However, binocular methods are usually computationally expensive and come with higher bandwidth costs. The stringent conditions mentioned above hinder the practical application of these methods. To circumvent these difficulties, increasing efforts have been made in developing cheaper 3D object detectors. They aim at detecting the 3D pose of objects from a single 3-channel RGB image~\cite{zhang2021objects,liu2022monocon,li2020rtm3d,Liu_2020_CVPR_Workshops,liu2021autoshape} that has been captured by a monocular camera.

Despite recent progress, the performance gap between monocular 3D object detectors and stereo-based or LiDAR-based detectors remains wide. Compared with stereo-based or LiDAR-based detectors, it is an ill-posed problem that recovers the depth information from a single image captured by a perspective camera. Due to the irreversible information loss introduced by the projection process of a perspective camera, the depth we can recover from a single image is theoretically very limited. Instead of directly regressing the object depth \cite{zhou2019objects,Liu_2020_CVPR_Workshops,luo2021m3dssd}, most recent works \cite{li2020rtm3d,zhang2021objects,liu2021autoshape} recover the depth of objects by modeling the relationship between the 3D geometric prior and the detected 2D keypoints. For instance, one can infer the depth of a target object by estimating the height of the 3D object and the height of the 2D projection, since the height of the 2D projection are inversely proportional to the depth when the 3D height is constant. These geometric constraint-based methods~\cite{zhang2021objects,li2020rtm3d,liu2021autoshape} usually show more strong generalization ability than the one that regresses the depth of objects directly~\cite{zhou2019objects,Liu_2020_CVPR_Workshops,luo2021m3dssd}. 

The depth estimation is particularly challenging when the object is occluded or truncated in the 2D view, as the surrounding 2D keypoints of the object cannot be fully recovered. Objects are only partially visible due to either the occlusion by other objects or the truncation due to the limited FoV (Field of View) of cameras. In this case, it is difficult to provide an accurate prediction for an object since its shape is largely missing. On the one hand, information that can be extracted from the partially visible object is very limited. In addition, the available occluded training samples are over-dominated by those fully-visible ones.  The training model is easily over-fit on the occluded objects. On the other hand, most of the existing object detection methods are center-based, for which an object center (instead of a 3D bounding box) is assigned to a detected object. However, for occluded/truncated objects, their center points often lie on other irrelevant objects. In such cases, the features derived from the center point and its neighboring area are mixed with irrelevant contents. 

In this paper, a novel auxiliary training task is designed to alleviate the over-fitting risk that is latent in the training process. Namely, an additional instance segmentation branch is added to the model. By learning to predict the visibility mask of objects, our model becomes ``shape-aware''. The feature is therefore derived from a real target object region, which prevents the model from overfitting to the noisy context surrounding the occluded objects.

However, the training task of instance segmentation requires pixel-level annotation on the monocular image, which is a laborious task to undertake. To overcome this difficulty, the pixel-level object annotation is approximated by utilizing the 3D object-level annotation and the corresponding point cloud.  We simply project the 3D points inside the 3D bounding box onto the image plane to form an instance mask used for training. Additionally, an uncertainty-weighted loss function is adopted to learn useful information from these noisy and sparse generated instance segmentation annotations.

In our paper, we also reveal the limitations of the current evaluation measure for 3D object detection. With the current IoU-based evaluation metrics, only the overlapping with the ground-truth objects is counted. The performance can be boosted by simply duplicating detected 3D bounding boxes and putting in different positions and layouts, which increases the chance of overlapping with the ground-truth while making no improvement in the 3D object detection. To address this issue, a new measure called average depth similarity (ADS) is proposed to evaluate the overlaps with the ground-truth in the object depth. In combination with 3D IoU-based metrics, a more objective evaluation can be made for 3D object detection methods that are based on the monocular image.

\section{Related Work}
\label{sec:related_work}
In this section, we briefly review the 3D object detection methods based on 3D point clouds and binocular images first.  Then the review focuses on the methods built upon a monocular image since our method also falls in this category.

In the representative LiDAR-based 3D object detection method~\cite{yang2018pixor}, 3D point clouds are encoded as 2D feature maps by projecting them into the bird-eye-view (BEV), then a 2D CNN is applied for detection. In VoxelNet~\cite{zhou2018voxelnet}, the neighboring cloud points are grouped into voxels. Thereafter, PointNet~\cite{qi2017pointnet} is adopted in each voxel for 
feature encoding. The 3D convolution, which is used in VoxelNet for object detection, comes with an expensive computational cost.  SECOND~\cite{yan2018second} improves the inference speed of VoxelNet by introducing sparse convolution based on the fact that many voxels are empty due to the point cloud sparsity. LiDAR-based detectors are able to provide accurate 3D detection results, however, the expensive sensor cost hinders the application of these methods. 

Besides 3D point clouds, depth information can also be recovered from stereo cameras. Stereo R-CNN~\cite{peiliang2019stereorcnn} 
extends the standard Faster R-CNN to detect and match the left and right objects, then refines the 3D detection with dense RoI (Region of Interest) alignment.  In~\cite{wang2019pseudo}, pseudo-LiDAR points extracted from the disparity map are directly used as the input for 3D object detectors, achieving state-of-the-art performance. Stereo camera-based methods usually require dense disparity estimation on the full image or each RoI, which is computationally expensive. They also introduce extra bandwidth and calibration requirements compared to monocular image-based methods.

Different from above methods, monocular based 3D object detectors only take a single RGB image as input. Amongst these methods, recent models are mostly center-based~\cite{zhou2019objects,zhang2021objects,liu2021autoshape,liu2022monocon}, where each object is represented with a representative center $\mathbf x_r=(u_c, v_c)$. Depth, dimensions, and orientation are estimated as attributes of the detected center point. Theoretically speaking, depth estimation from a monocular image is ill-posed. There are various ways to undertake depth estimation. They can be roughly divided into three categories, namely the direct depth regression, depth-map-based, and geometric constraint-based methods. Direct depth regression methods~\cite{zhou2019objects,Liu_2020_CVPR_Workshops,luo2021m3dssd} regress the depth as the 3D attributes of the objects directly. These direct depth regression methods are efficient since they do not rely on any additional depth-map estimation modules. 
The depth map-based methods \cite{simonelli2021we,wang2019pseudo} are also known as the pseudo-LiDAR (PL) based method, employs an off-the-shelf dense depth map estimation Convolutional Neural Network (CNN) as a substitute for LiDAR sensors. With the resulting pseudo-LiDAR, the detection task is usually achieved by applying a state-of-the-art LiDAR-based detector. Due to biased training protocol~\cite{simonelli2021we}, superior performance for PL-based methods is only observed on the validation set. According to \cite{simonelli2021we}, the training set adopted by their depth estimation modules heavily overlaps with the \textit{val} set of the PL-based detectors.

\begin{figure}
  \centering
  \includegraphics[width=0.8\linewidth]{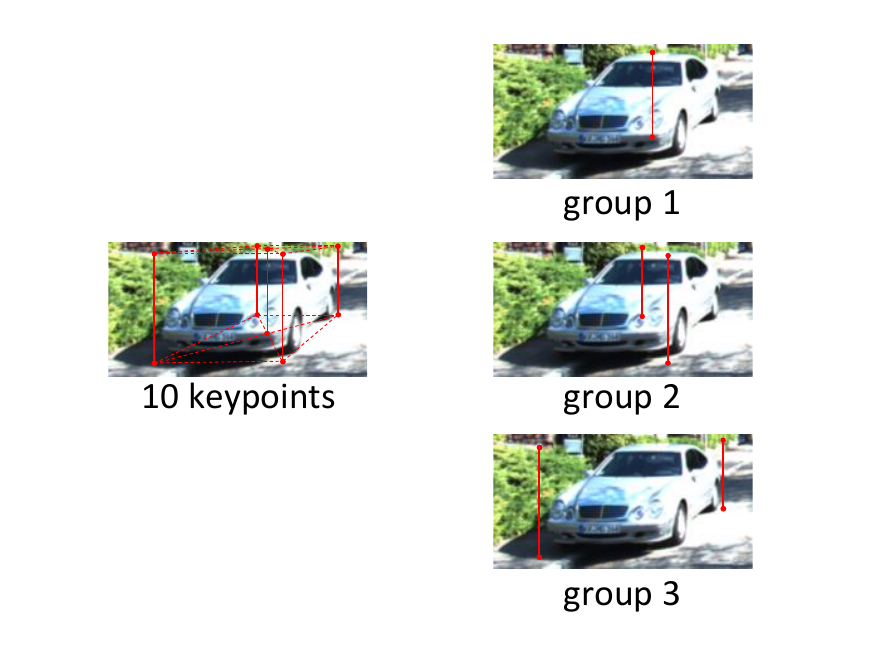}
  \caption{
    Five vertical lines can be extracted from 10 keypoints. These 
    lines are divided into three groups, each of which can produce an 
    independent depth prediction. \label{fig:keypoint_definition}
  }
\end{figure}

Apart from the above two types of depth estimation, the task can also be implemented by leveraging 3D-2D geometric constraints. RTM3D \cite{li2020rtm3d} estimates nine predefined keypoints which introduce \textit{18} projection constraints (\textit{9} pairs of x and y) for each object. Then the depth of each object can be recovered by solving a nonlinear least-squares optimization problem. Due to the variety of car layouts, the estimated x-coordinate is usually less reliable than the y-coordinate for one keypoint. Therefore, only the geometric constraint on the height of keypoints is utilized in Monoflex~\cite{zhang2021objects}. In addition, it also divides \textit{10} keypoints into three groups (shown in Figure~\ref{fig:keypoint_definition}) to produce three independent depth predictions. The final depth is a weighted average of all the depth predictions.

In the above geometric-constraint-based methods, objects are modeled as 3D bounding boxes, and keypoints are often defined as 
corners, centers, or top/bottom points of the bounding boxes. These keypoints mostly lie on the background, \textit{e.g.}, the sky or the ground. Deep MANTA \cite{chabot2017deepmanta} and AutoShape \cite{liu2021autoshape} address this issue by annotating the keypoints of the training data with CAD templates via a semi-autonomous/autonomous approach. However, due to the mismatch between the predefined CAD model and the real object, the improvement that these methods achieve is very limited. In this paper, an auxiliary training task is introduced. Namely, instance segmentation is incorporated as another task for the detection network. As the task itself is shape-aware, the detected object regions are expected to shrink from a bounding box to the real object area. Moreover, since the annotations on the 3D point cloud are wisely utilized to produce the segmentation mask, the introduced segmentation head boosts the overall detection performance while no laborious pixel-level annotation is required.
\section{Shape-aware 3D Object Detection}
\label{sec:proposed_method}
As aforementioned, the existing monocular 3D object detection models easily get over-fit on the occluded objects. First of all, in the monocular detectors, objects are modeled as representative centers. However, for the occluded objects, their center may lie on other objects, leading to the polluted feature representation that is derived from the center. Moreover, it is challenging for the network to recover the 3D properties of an object when only a small area of the target object is visible. Furthermore, very few samples are available for the severely occluded objects in the training set. 

\subsection {Shape-aware Auxiliary Training Task}
To address the above issue, we aim to train a model that is able to estimate the shape of a target object. Given the precise shape of a detected object is known, the detector would be able to focus on the real visible areas of an object. Therefore, more precise detection of occluded objects could be expected.

\begin{figure}
  \centering 
  \includegraphics[width=8.5cm]{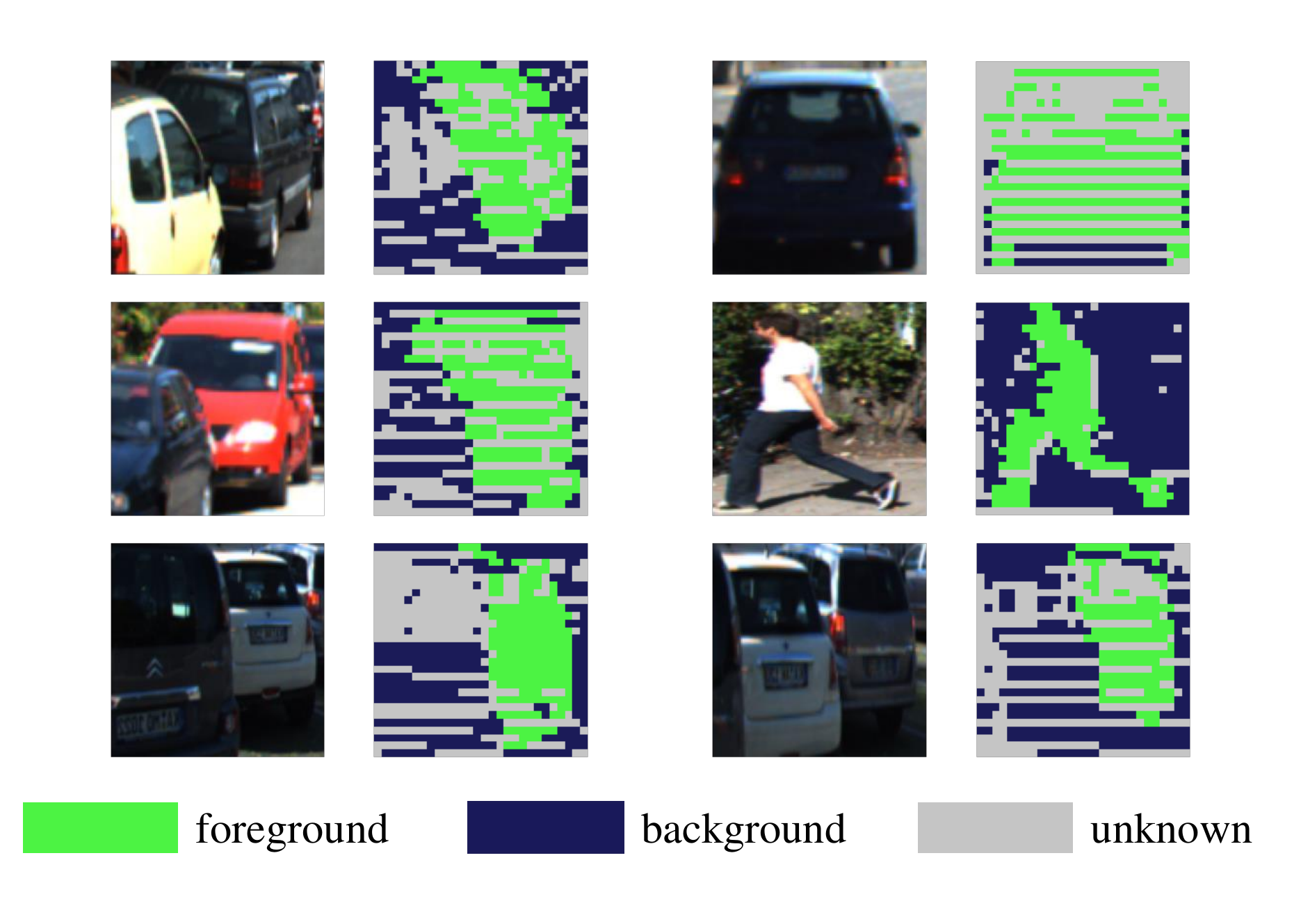}
  \caption{Instance level masks generated by projecting LiDAR points onto the image plane and treating points inside the 3D bounding box as foreground.\label{fig:instance_mask}}
\end{figure}


\begin{figure}
  \centering 
  \includegraphics[width=8.5cm]{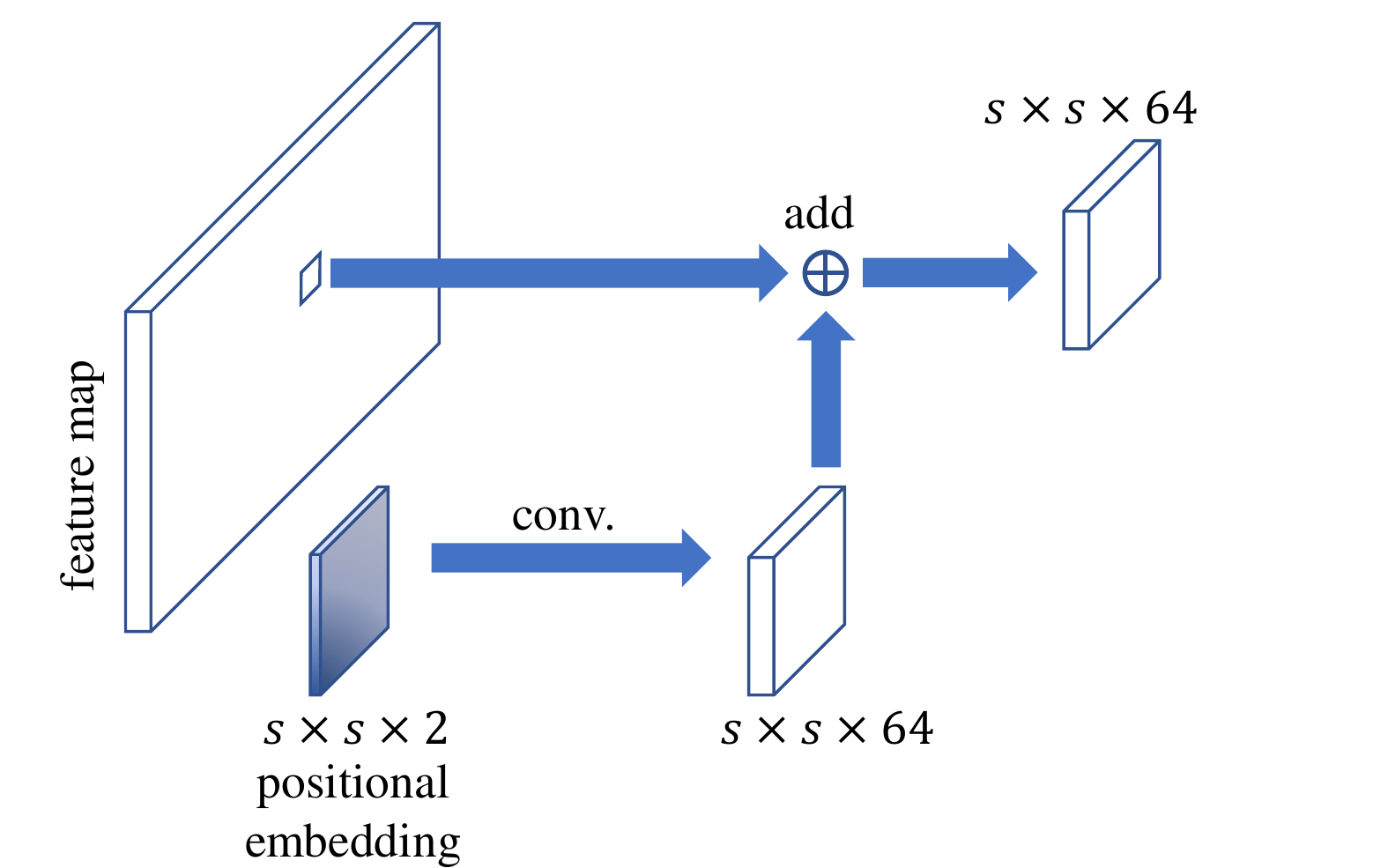}  
  \caption{The feature expanding module. In this module, the feature vector extracted from the representative center is expanded to a square feature map. This feature map is further fed into subsequent CNN modules to produce the final mask prediction.
     \label{fig:feature_expanding}}
\end{figure}

Attaching to the backbone network, a center-based segmentation branch is added, which is parallel to the detection head. Since the object detection pipeline is also center-based, these two tasks are compatible to each other. As illustrated in Figure~\ref{fig:feature_expanding}, we extract a feature vector $\mathbf f_i \in \mathbb R^{64}$ for the $i$th object, then it's added with a predefined position embedding map $\mathbf f_{pos} \in \mathbb R^{s, s, 64}$ with broadcasting. A 
feature map with size $s\times s\times 64$ is, therefore, produced. The position embedding map $\mathbf f_{pos}$ is a predefined image with two channels. Each pixel at position $(x_p, y_p)$ of $\mathbf f_{pos}$ equals to a linear transformed coordinate $(\tilde x_{p}, \tilde y_{p}) = \left( \frac{x_p}{s} - 0.5, \frac{y_p}{s} - 0.5 \right)$.

In order to train this center-based segmentation branch, the instance masks must be ready for the training images. However, pixel-level 
annotation is expensive. To circumvent this difficulty, we alternatively extract the instance masks from the 3D bounding boxes and the original point cloud. Inspired by the ground-truth sampling strategy used by LiDAR-based 3D detection methods~\cite{yan2018second,zhu2019class}, we treat points inside the bounding boxes as instance-level foreground 
points. When the whole cloud of points are projected onto the image plane, the pixel-level mask label is produced for each object by checking whether a point is inside the box or not. An $s\times s$ square image is used to represent the ground-truth mask. 
If there are two or more points that fall into the same pixel of the image, we keep one of them at random. 
As illustrated in Figure~\ref{fig:instance_mask}, the generated ground-truth masks are coarse and sparse, however, they are sufficient to make the whole detection pipeline become aware of the object shapes.

To alleviate the interference from the noisy data, an uncertainty weighted cross-entropy loss is designed for the instance segmentation head

\begin{equation}
\begin{aligned}
L_{seg} =& -\frac{1}{s^2\sigma_{seg}^2} \sum_{(r, c)\in s\times s} (L_{seg}^+(r,c) + L_{seg}^-(r, c))\\
                         & + \frac{\log\sigma_{seg}}{s^2} \sum_{(r, c) \in s\times s}\mathbf 1(y_{rc} \neq -1) 
\end{aligned},
\label{eqn:uncertainty_loss}
\end{equation}
where 
\begin{equation}
\nonumber
\begin{aligned}
L_{seg}^+(r, c) &= \mathbf 1(y_{rc} = 1) \log p_{rc}, \\
L_{seg}^-(r, c) &= \mathbf 1(y_{rc} = 0) \log(1 - p_{rc}).     \\
\end{aligned}
\end{equation}

In Eqn.~\ref{eqn:uncertainty_loss}, the indicator function $\mathbf 1(\cdot)$ equals to \textit{1} if the 
input expression is true, and \textit{0} otherwise. The label is $y_{rc} \in \left\{1, 0, -1\right\}$ when the pixel 
is foreground, background or unknown respectively. The uncertainty value $\sigma_{seg}$ is a learnable parameter, that indicates how much 
noise we have in this task. In practice, we train the model to predict the $\log\sigma_{seg}$ to avoid division by zero.

In order to increase the number of training samples as well as make the model more robust to the position error of the detected center,  we assign the instance labels to each point that is inside the $3\times3$ neighborhood of the object center point. Namely, we first collect all points that belong to the $3\times 3$ neighborhood of representative points. These points share the same object label as the center closest to them in terms of  \textit{Euclidean} distance.  The center sampling is adopted in the regression heads and the segmentation head. 
For regression heads, the center sampling operation is performed on regressed attributes. For the segmentation head, the center sampling is applied on original features and before the segmentation head to avoid producing abundant masks on background positions. The center sampling strategy does not affect the heatmap prediction. We still apply the penalty-reduced focal loss \cite{lin2017focal,zhou2019objects,law2018cornernet} to train the heatmap. At the test time, we can simply extract the object 
predictions by finding the maximum points of the heatmap through a max-pooling operation.

For efficiency, the segmentation head is turned on only during the training. It is already sufficient since the instance segmentation training is designed to assist the training of detection and regression tasks. 

\subsection{Multi-Task Learning in 3D Object Detection} 
Similar to other center-based methods~\cite{zhou2019objects,zhou2021probabilistic}, each target object is detected as a center point along with its attributes in our method. The detection is built upon the DLA-34 backbone network. There are several target parameters to be estimated (listed in Table~\ref{tab:heads_and_parameters}) in our model. In general, it is a multi-task deep learning framework. The classification head estimates the raw location $(u_f, v_f)$ of the 2D center point. Other attributes are regressed by several regression heads. The segmentation head is designed to boost the performance of other heads.

\begin{table}[htbp]
  \centering
  \caption{Detection heads and the corresponding parameters to be estimated}
    \begin{tabular}{l|l|l}
    \toprule
    Head                           & Task        & Target Paras. \\
    \midrule
    Classification                 & Heatmap           & $u_f, v_f$ \\
    \hline
    Segmentation                   & Shape-Aware Feat. & - \\
    \hline
    \multirow{7}[0]{*}{Regression} & Depth             & $z$ \\
                                   & Orientation       & $r_y$ \\
                                   & 2D Bounding Box   & $l_b, t_b, r_b, b_b$ \\
                                   & 3D Dimensions     & $h, w, l$ \\
                                   & Center Offset     & $\delta_u, \delta_v$ \\
                                   & Keypoints         & $\mathbf x_{kpt}$ \\
                                   & Uncertainties     & $\sigma_1, \sigma_2, \sigma_3, \sigma_4$ \\
    \bottomrule
    \end{tabular}%
  \label{tab:heads_and_parameters}%
\end{table}%

\textbf{2D Detection} Similar to FCOS \cite{tian2019fcos}, the 2D bounding box is represented as distances from the representative center to four sides of the box. GIoU loss \cite{Rezatofighi_2019_CVPR} is adopted to learn 2D bounding boxes since it is robust to object scale changes.

\textbf{3D Detection} As illustrated in Figure~\ref{fig:keypoint_definition}, the keypoint regression head predicts \textit{10} keypoints for each object, from which we can extract five vertical lines. The depth value of vertical line $l$ is estimated by utilizing the relative proportion between pixel height and estimated object height \cite{zhang2021objects,cai2020monocular} 
\begin{equation}
z_l=\frac{f\times h}{h_l},
\end{equation}
where $f$ is the focal length. $h_l$ and $h$ are the height of line $l$ in pixel and the 3D height of the object respectively. Following with~\cite{zhang2021objects}, five vertical lines are divided into three groups, and the depth value of each vertical line is averaged within each group, resulting in three independent center depth values. In addition, another depth value is also estimated directly by regression. The resulting \textit{4} depth values are weighted by their uncertainties (given in Eqn.~\ref{eqn:fdepth}).
\begin{equation}
z = \left(\sum_{i=1}^{4}\frac{z_i}{\sigma_i}\right)/\left(\sum_{i=1}^{4}\frac{1}{\sigma_i}\right),
\label{eqn:fdepth}
\end{equation}
where the uncertainty estimation branch is trained under the laplacian uncertainty loss~\cite{kendall2019geometry,chen2021monorun,kendall2017uncertainties,kendall2018multi}.

When an object is detected on the heatmap at position $(u_f, v_f)$, its center $\mathbf x_c$ is given by $(s_0 {\cdot} u_f + \delta_u, s_0 {\cdot} v_f + \delta_v)$, where $s_0$ is the downsampling factor. With the predicted projected 3D center 
$\mathbf x_c = (u_c, v_c)$, The object location can be decoded as 
\begin{equation}
(x, y, z) = \left(\frac{(u_c - c_u) {\cdot} z}{f}, \frac{(v_c - c_v) {\cdot} z}{f}, z\right),
\end{equation}
where $(c_u, c_v)$ is the principle point. 

For dimension estimation, the relative changes with 
respect to the average dimension is predicted. For each 
class $c$, the average dimension is denoted as 
$(\bar h_c, \bar w_c, \bar l_c)$, the L1 loss for dimension 
regress is defined as
\begin{equation}
L_{dim} = \sum_{k\in{h, w, l}} \left| \bar k_c {\cdot} e^{\delta_k} - k^* \right|,
\end{equation}
where $k^*$ is the ground-truth dimension, $\delta_k$ is the relative offset to be regressed. The dimensions are estimated by scaling the average dimensions, namely $(h, w, l) = (\bar h_c {\cdot} e^{\delta_h}, \bar w_c {\cdot} e^{\delta_w}, \bar l_c {\cdot} e^{\delta_l})$.

MultiBin loss \cite{mousavian20173d} is used to estimate the local 
orientation $\alpha$. The global 
orientation $r_y$ can be obtained by calculating
\begin{equation}
r_y = \alpha + \arctan(x/z).
\end{equation}

The final 3D bounding box is then encoded as $(x, y, z, w, h, l, r_y)$. 
We refer readers to \cite{zhang2021objects} for more details of the 3D 
detection head.

\subsection{Implementation Details}
Following~\cite{Liu_2020_CVPR_Workshops,zhang2021objects,chen2020monopair}, a modified DLA-34~\cite{zhou2019objects} is adopted as our backbone. For each input image\footnote{All the input images are padded into the same size of $384\times 1280$.}, the backbone produces a feature map with the down-sampling ratio \textit{4}. Different headers are responsible to estimated different parameters. The parameters to be estimated and the corresponding header are shown in Table~\ref{tab:heads_and_parameters}. Every regression head consists of one $3 \times 3 \times 256$ convolution layer, BatchNorm \cite{ioffe2015batch}, ReLU, and another $1 \times 1 \times c_o$ convolution layer, where $c_o$ is the output channels. For heatmap prediction, we use the same structure except that there is a sigmoid function padded after the final Conv layer. For center offset and heatmap prediction, the edge fusion~\cite{zhang2021objects} module is applied after the $3{\times}3$ Conv layer to  decouple the feature learning of truncated objects. For the instance segmentation head, a two-layer CNN block with a ReLU activation function is used to process the predefined position embedding map. The processed position embedding map is then added with extracted features vectors with broadcasting. The resulted tensor is then fed into another two-layer CNN, which has a GroupNorm \cite{wu2018group} layer embedded before each ReLU. The final mask is obtained by further applying one convolution layer with a sigmoid activation function. 

The model is trained using AdamW optimizer with an initial learning rate as $3\times 10^{-4}$ and decay rate as 
$1\times 10^{-5}$. We train the model for \textit{200} epochs with a batch size of \textit{18} on three 1080-Ti GPUs. The learning rate is divided by \textit{10} at the \textit{190th} epoch and the \textit{195th} epoch.

\section{Average Depth Similarity}
In the literature, KITTI~\cite{Geiger2012CVPR,Geiger2013IJRR} is one of the most popularly used evaluation benchmarks. The performance of monocular 3D object detection is measured by the 3D IoU-based Average Precision (AP) on the 3D space (${\text{AP}_{3D}}$) and the BEV (${\text{AP}_{BEV}}$).
A detection is considered positive if it overlaps a ground-truth with an IoU larger than a threshold. Such metrics are reasonable to  
evaluate LiDAR-based or stereo-based methods. However, we find they could not reflect the real performance of a monocular object detector. 

\begin{figure}
  \begin{minipage}[b]{1.0\linewidth}
    \centering
    \centerline{\includegraphics[width=8.5cm]{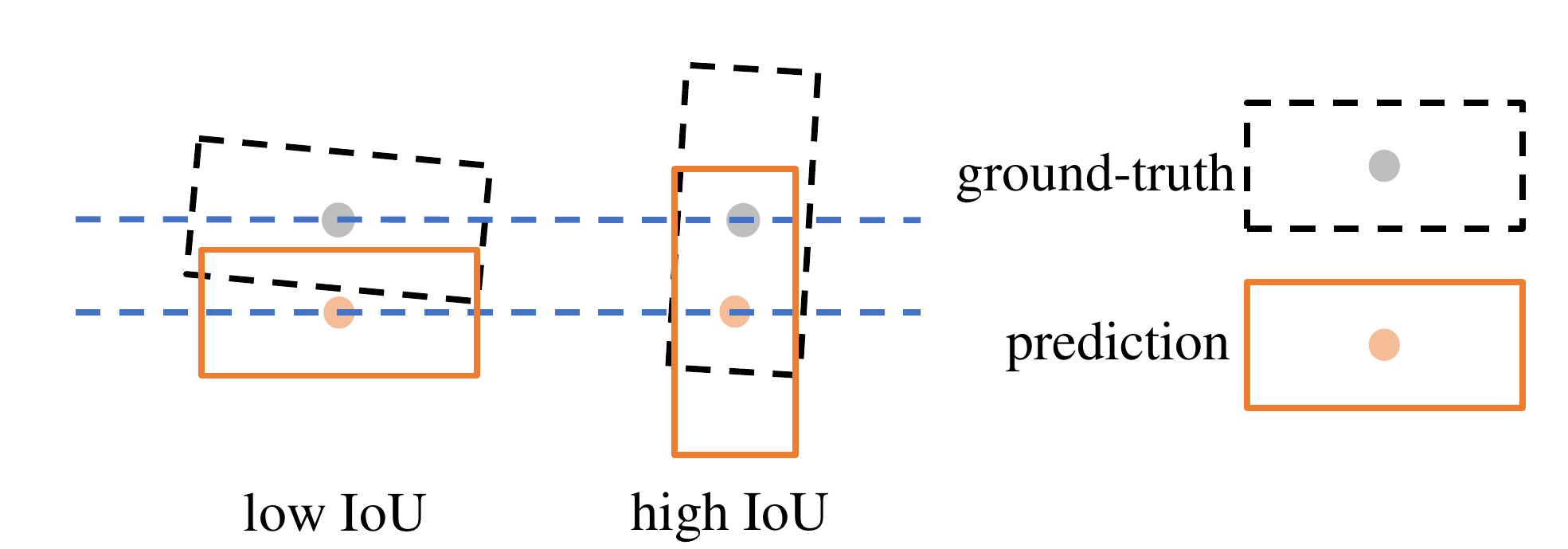}}
    \centerline{(a) Vertical objects produce higher IoU}\medskip
  \end{minipage}
  \begin{minipage}[b]{1.0\linewidth}
    \centering
    \centerline{\includegraphics[width=8.5cm]{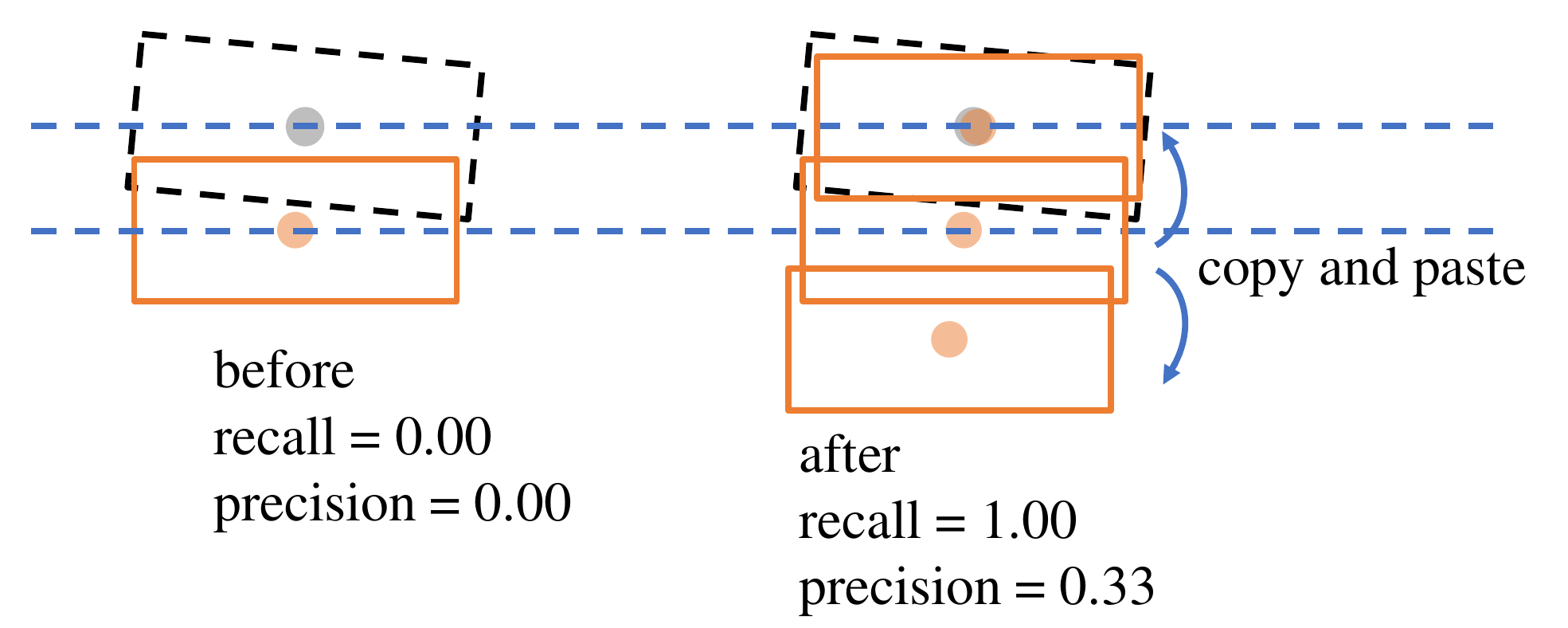}}
    \centerline{(b) The copy-and-paste strategy}\medskip
  \end{minipage}
  \caption{Two scenarios where  ${\text{AP}_{3D}}$  cannot reflect the real performance of a monocular 3D detector.}
  \label{fig:drawbacks_of_ap}
\end{figure}

First of all,  as pointed out in \cite{ma2021delving}, the depth estimation error increases with respect to the object's distance to the camera. The detected objects that are far away from the camera are largely ignored due to low 3D IoU. Figure~\ref{fig:drawbacks_of_ap} further shows another two scenarios in which we cannot have an honest observation about the detector with 3D IoU-based evaluation. In Figure~\ref{fig:drawbacks_of_ap}(a), we can see the evaluation favors object in vertical layout even both of them are estimated with the same depth accuracy.  In Figure~\ref{fig:drawbacks_of_ap}(b), we show a trick that ${\text{AP}_{3D}}$ can be cheated. ${\text{AP}_{3D}}$ score is boosted by simply duplicating and shifting several distant predictions. As we will see later in the experiment, ${\text{AP}_{3D}}$ can be boosted by \textit{25\%} by such simple ``result sampling''. Although this trick is effective in producing better ${\text{AP}_{3D}}$, it is essentially a trick that makes a better trade-off between precision and recall. It is non-pragmatic since it neither improves the quality of the results nor enhances the ability of the model.

To address this issue, average depth similarity (ADS) is proposed as a complementary to ${\text{AP}_{3D}}$.
Compared to ${\text{AP}_{3D}}$ which is based on 3D-IoU, ADS matches the 2D predictions with the ground-truth. The depth similarity is defined as
\begin{equation}
s_d(r) = \frac{1}{\left| \mathcal D (r) \right|} \sum_{i\in \mathcal D(r)} \exp(-\left|\Delta_d^{(i)}\right|) \delta_i,
\label{eqn:ads}
\end{equation}
where $\mathcal D(r)$ are the set of all detected objects at recall rate $r$ and $\Delta_d^{i}$ is the difference of the depth of detection $i$. $\delta_i$ is \textit{1} if detection $i$ has been associated with a ground-truth object and \textit{0} 
otherwise. The definition of ADS follows with the original design of AOS \cite{Geiger2012CVPR,Geiger2013IJRR},
while the difference lies in the similarity definition. The error in depth $\Delta_d^{(i)} \in [0, +\infty)$ is normalized into the range of $[0,~1]$. 

\begin{figure}
  \centering
  \includegraphics[width=0.65\linewidth]{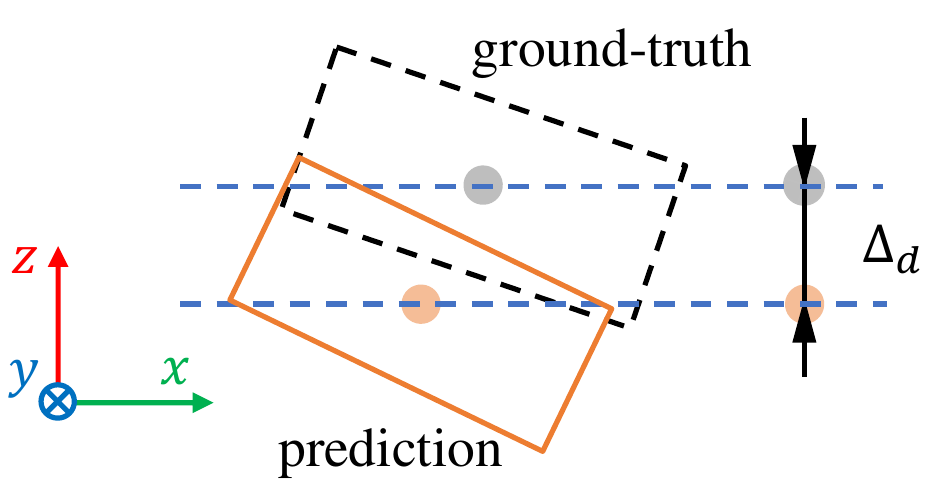}
  \caption{The depth similarity is measured by calculating the absolute distance error along the z-axis.\label{fig:ads}}
\end{figure}

As illustrated in Figure~\ref{fig:ads}, the calculation of depth similarity solely depends on the absolute depth error along the z-axis. Unlike the 3D IoU, the calculation of depth similarity is independent of object attributes such as size, orientation, and position. In addition, ADS avoids penalizing predictions with small IoUs. Therefore, it enables a comprehensive observation of the behavior of a monocular detector  when combined with traditional evaluation metrics ${\text{AP}_{3D}}$ and ${\text{AP}_{BEV}}$. 

\section{Experiments}
\label{sec:experiments}
\begin{table*}[h!]
  \caption{Ablation study on the KITTI \textit{val} set for three categories, \textit{i.e.}, the Car, Pedestrian and Cyclist}
  \label{tab:ablations_3class}
  \resizebox{\textwidth}{!}{%
      \begin{tabular}{l|cc|cc|cc|cc|cc|cc|cc|cc|cc}
      \toprule
      \multirow{3}{*}{Methods}    & \multicolumn{6}{c|}{Car}                   & \multicolumn{6}{c|}{Pedestrian} & \multicolumn{6}{c}{Cyclist}\\ 
      \cline{2-19}
                                  & \multicolumn{2}{c|}{Easy}            & \multicolumn{2}{c|}{Moderate}         & \multicolumn{2}{c|}{Hard}           & \multicolumn{2}{c|}{Easy}          & \multicolumn{2}{c|}{Moderate}       & \multicolumn{2}{c|}{Hard}            & \multicolumn{2}{c|}{Easy}           & \multicolumn{2}{c|}{Moderate}       & \multicolumn{2}{c}{Hard} \\
                                  & ${\text{AP}_{3D}}$ & ${\text{ADS}}$  & ${\text{AP}_{3D}}$ & ${\text{ADS}}$   & ${\text{AP}_{3D}}$ & ${\text{ADS}}$ &${\text{AP}_{3D}}$ & ${\text{ADS}}$ & ${\text{AP}_{3D}}$ & ${\text{ADS}}$ & ${\text{AP}_{3D}}$  & ${\text{ADS}}$ & ${\text{AP}_{3D}}$ & ${\text{ADS}}$ & ${\text{AP}_{3D}}$ & ${\text{ADS}}$ & ${\text{AP}_{3D}}$ & ${\text{ADS}}$ \\ 
      \midrule
      MonoFlex*                   & 21.74              & 69.59           & 15.79                & 60.16          & 13.32              & 53.02          & 5.47              & 46.23          & 4.40          & 36.84               & 3.47                & 31.12          & 5.24               & 36.41          & 2.69          & 21.28               & 2.59               & 19.94 \\
      + CS                        & 23.25              & 70.00           & 16.78                & 60.79          & 14.81              & 54.96          & 7.54              & 52.37          & 5.54          & 43.77               & 4.52                & 36.45          & 5.62               & 37.82          & 3.17          & 23.57               & 2.72               & 22.15 \\
      + CS + SA                   & \textbf{24.92}     & \textbf{73.53}  & \textbf{18.39}       & \textbf{62.68} & \textbf{15.56}     & \textbf{56.61} & \textbf{8.80}     & \textbf{53.34} & \textbf{6.88} & \textbf{44.56}      & \textbf{5.78}       & \textbf{37.13} & \textbf{6.77}      & \textbf{44.18} & \textbf{3.24} & \textbf{26.86}      & \textbf{3.12}      & \textbf{24.93} \\
      \bottomrule
      \end{tabular}
  }
\end{table*}

\begin{table*}[ht!]
  \centering
  \caption{
    Comparisons for Car category on the KITTI benchmark. 
    For models marked with *, we evaluate their performance 
    on the \textit{val} set by training with official codes. 
    For other models, their results on the \textit{val} set are 
     either cited from their papers or by evaluating the released pre-trained model from the authors.
    Methods are ordered by their ${\text{AP}_{3D}}$ values under the moderate setting.
    For clarity, the best results are in {\color{red}red}, while the second is highlighted in {\color{blue}blue}
  }
    \begin{tabular}{l|l|l|lll|lll}
    \toprule
    \multicolumn{1}{c|}{\multirow{2}[0]{*}{Name}} & \multicolumn{1}{c|}{\multirow{2}[0]{*}{Publication}}    & \multirow{2}[0]{*}{Extra Dataset} & \multicolumn{3}{c|}{$\text{AP}_{3D}$ (test/val)}                                                                          & \multicolumn{3}{c}{${\text{ADS}}$ (val)} \\
                                                  &                &                                   & Easy                                   & Moderate                               & Hard                                    & Easy                & Moderate                 & Hard \\
    \midrule
    AutoShape \cite{liu2021autoshape}             & ICCV 2021      & KINS                              & 22.47/20.09                            & 14.17/14.65                            & 11.36/12.07                             & -                   & -                        & -  \\
    LPCG-M3D-RPN \cite{peng2021lidar}             & arxiv preprint & KITTI raw                         & 22.73/26.17                            & 14.82/19.61                            & 12.88/16.80                             & -                   & -                        & -  \\
    DD3D \cite{park2021pseudo}                    & ICCV 2021      & DDAD15M                           & 23.22/ -                               & 16.34/ -                               & 14.20/ -                                & -                   & -                        & -  \\
    LPCG-MonoFlex \cite{peng2021lidar}            & arxiv preprint & KITTI raw                         & 25.56/31.15                            & 17.80/23.42                            & 15.38/20.60                             & -                   & -                        & -  \\
    \hline
    MonoRUn \cite{chen2021monorun}                & CVPR 2021      & \multirow{7}[0]{*}{None}          & 19.65/21.63                            & 12.30/15.22                            & 10.58/12.83                             & 67.95               & 58.91                    & 52.94  \\
    Ground-Aware* \cite{liu2021ground}            & IEEE RA-L 2021 &                                   & 21.65/23.14                            & 13.25/15.71                            & 9.91/11.76                              & 69.42               & 54.47                    & 42.51  \\
    MonoFlex \cite{zhang2021objects}              & CVPR 2021      &                                   & 19.94/24.22                            & 13.89/17.34                            & 12.07/15.14                             & 71.59               & {\color{blue}62.19}      & {\color{blue}56.10}\\
    DLE* \cite{liu2021deep}                       & BMVC 2021      &                                   & {\color{red}24.23}/{\color{blue}25.58} & 14.33/16.50                            & 10.30/12.27                             & 70.26               & 56.44                    & 43.01  \\
    GUPNet \cite{lu2021geometry}                  & ICCV 2021      &                                   & 22.26/23.19                            & 15.02/16.23                            & 13.12/13.57                             & 71.17               & 59.32                    & 52.14  \\
    MonoCon \cite{liu2022monocon}                & AAAI 2022      &                                   & 22.50/{\color{red}26.33}               & {\color{blue}16.46}/{\color{red}19.03} & {\color{red}13.95}/{\color{red}16.00}   & {\color{blue}72.15} & 60.29                    & 54.40  \\
    \hline 
    Ours                                          & -              & None                              & {\color{blue}23.84}/24.92              & {\color{red}16.52}/{\color{blue}18.39} & {\color{blue}13.88}/{\color{blue}15.56} & {\color{red}73.53}  & {\color{red}62.68}       & {\color{red}56.61}  \\
    \bottomrule
    \end{tabular}%
  \label{tab:performance}%
\end{table*}%

Our method is evaluated on the popular KITTI dataset~\cite{Geiger2013IJRR, Geiger2012CVPR} in comparison to state-of-the-art methods. The dataset consists of \textit{7,481} training images and \textit{7,518} testing images. Due to the restrictions on the \textit{test} set submission to the KITTI official site, the training images are split into \textit{train} (3,712) and \textit{val} (3,769) sets following~\cite{chen20153d}. All our models jointly predict three categories, including Car, Pedestrian, and Cyclist.

The methods we consider in the experiment include one-stage center-based methods such as AutoShape~\cite{liu2021autoshape}, LPCG-MonoFlex~\cite{peng2021lidar}, MonoCon~\cite{liu2022monocon}, MonoFlex~\cite{zhang2021objects} and GUPNet~\cite{kendall2019geometry}. The comparison also covers one-stage anchor-based methods such as Ground-Aware~\cite{liu2021ground}, DLE~\cite{liu2021deep}, and LPCG-M3D-RPN~\cite{peng2021lidar} and one-stage FCOS-like method DD3D~\cite{park2021pseudo}. A two-stage method MonoRUn~\cite{chen2021monorun} is incorporated in our comparative study as well. In the evaluation, MonoFlex~\cite{zhang2021objects} that is retrained with the same settings as our method is treated as the comparison baseline. For clarity, this run is given as ``MonoFlex*'', while the run that is loyal to the original paper is given as ``MonoFlex''.

\begin{table}[h!]
  \centering
  \caption{Validation Test on ADS Measure}
  \label{tab:result_sampling}
      \begin{tabular}{l|lll|lll}
      \toprule
                                 & \multicolumn{3}{c|}{${\text{AP}_{3D}}$} & \multicolumn{3}{c}{ADS}                                                                    \\
      \multirow{-2}{*}{Methods}  & Easy           & Mod.           & Hard             & Easy   & Mod. & Hard   \\
      \midrule
      Ours               & 24.92          & 18.39          & 15.56            & \textbf{73.53} & \textbf{62.68} & \textbf{56.61}  \\
      + sampling                & \textbf{27.86} & \textbf{22.31} & \textbf{19.42}   & 69.22          & 52.29    & 46.45           \\
      \bottomrule
      \end{tabular}
\end{table}

Before we proceed with the comprehensive empirical study, the proposed evaluation measure, namely average depth similarity (ADS) is validated. 
A simple sampling~\cite{peng2022digging} is conducted on the results produced by our method. As shown in Table~\ref{tab:result_sampling}, the $\text{AP}_{3D}$ score of our method becomes considerably higher with simple sampling. This is where ${\text{ADS}}$ comes to complement. The performance of ``+ sampling'' run is much poorer when it is measured by ${\text{ADS}}$. This is because the $\text{AP}_{3D}$ of the ``+ sampling'' increases at the cost of producing more blind predictions. It, therefore, makes no contribution to the quality of 2D bounding boxes nor depth predictions. In the following experiments, both $\text{AP}_{3D}$ and ${\text{ADS}}$ scores are reported on the \textit{val} for all the methods.

\subsection{Ablation Study}\label{sec:ablation}
In our first experiment, an ablation study is conducted to show the contributions of center sampling (CS) and shape-aware training (SA), both of which are introduced by us. We observed that, when applying center sampling alone, \textit{6.27\%} and \textit{1.05\%} improvement in terms of ${\text{AP}_{3D}}$ and ${\text{ADS}}$ respectively are observed on the moderate objects. When the shape-aware
auxiliary training head is further integrated, extra \textit{9.59\%} and \textit{3.11\%} respectively are observed. As shown in Table~\ref{tab:ablations_3class}, this improvement is consistent on the Pedestrian and Cyclist categories.

\begin{table}[h!]
  \centering
  \caption{Comparison between different RoI feature extractors. Experiments are conducted on the KITTI \textit{val} set}
  \label{tab:roi_feature_extractors}
      \begin{tabular}{l|rrr|rrr}
      \toprule
                                               & \multicolumn{3}{c|}{${\text{AP}_{3D}}$} & \multicolumn{3}{c}{${\text{ADS}}$}                                                                    \\
      \multirow{-2}{*}{Methods} & \multicolumn{1}{c}{Easy} & Mod.      & Hard  & Easy   & Mod. & Hard    \\
      \midrule
      MonoFlex*                    & 21.74          & 15.79          & 13.32          & 69.59          & 60.16          & 53.02  \\
      + RoIAlign                   & 19.97          & 14.65	         & 12.91          & 69.19	         & 59.79	        & 52.86  \\
      + Ours                       & \textbf{24.92} & \textbf{18.39} & \textbf{15.56} & \textbf{73.53} & \textbf{62.68} & \textbf{56.61}  \\
      \bottomrule
      \end{tabular}
\end{table}

In our second experiment, we further confirm the choice of feature expanding (illustrated in Figure~\ref{fig:feature_expanding}) as the feature extractor in the segmentation head. In this ablation study, the performance of using feature expanding is compared to the configuration that uses the RoIAlign~\cite{he2018mask} in the segmentation head. As shown in  Table~\ref{tab:roi_feature_extractors}, the performance of the detector drops when RoIAlign is used as the feature extractor in the segmentation head. In contrast, considerable improvement (in comparison to MonoFlex*) is observed when the RoIAlign is replaced with feature expanding in the feature extractor. As discussed earlier, the misalignment between the RoI-based and center-based representation leads to potential conflicts between two training heads. The segmentation becomes harmful to the detection task, which leads to even poorer performance than the case it is, otherwise, not integrated.

\subsection{Performance Analysis}
\begin{figure*}
  \centering
  \includegraphics[align=c,width=6.3cm]{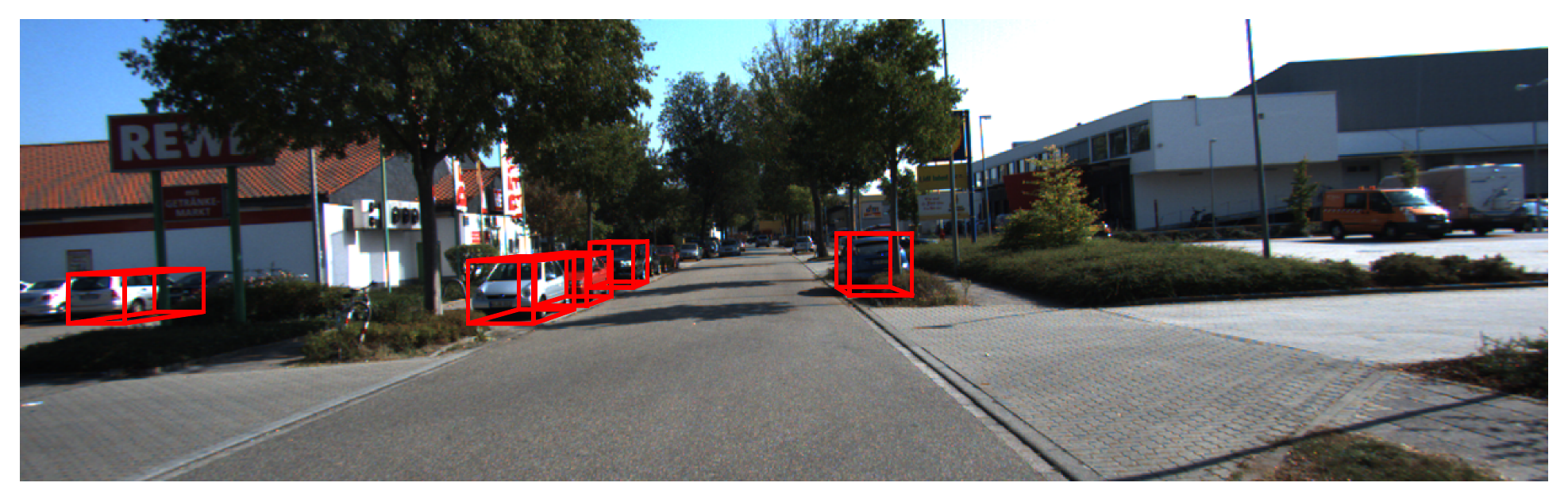}
  \includegraphics[align=c,width=6.3cm]{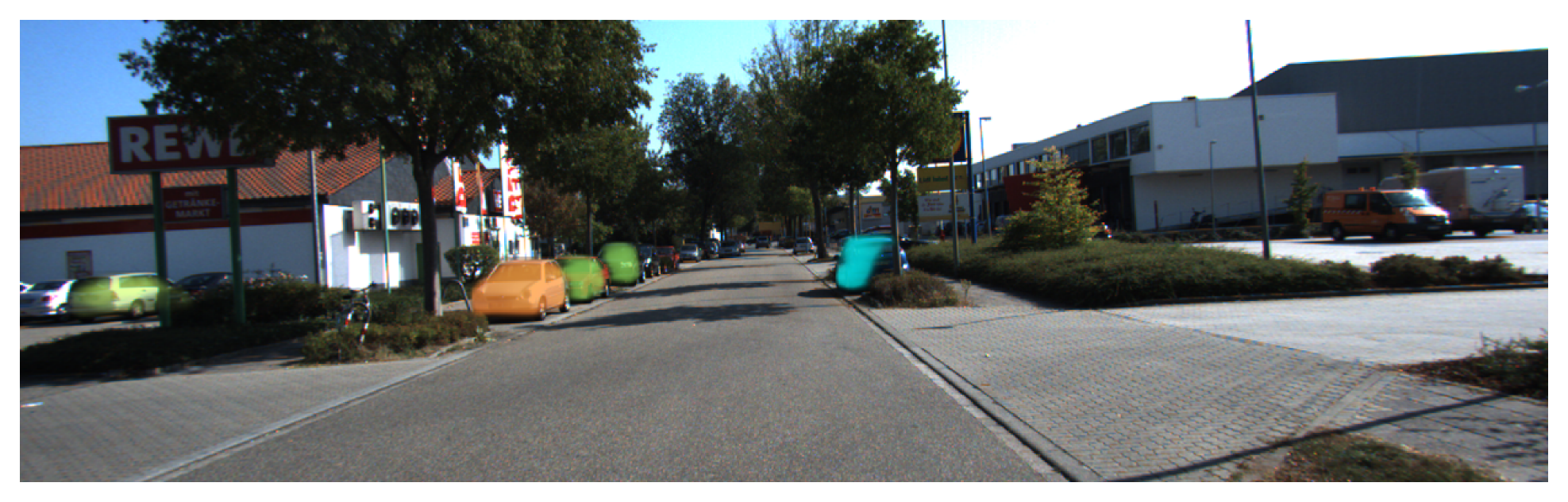}
  \raisebox{-0.18\height}{\includegraphics[align=c,width=5.0cm]{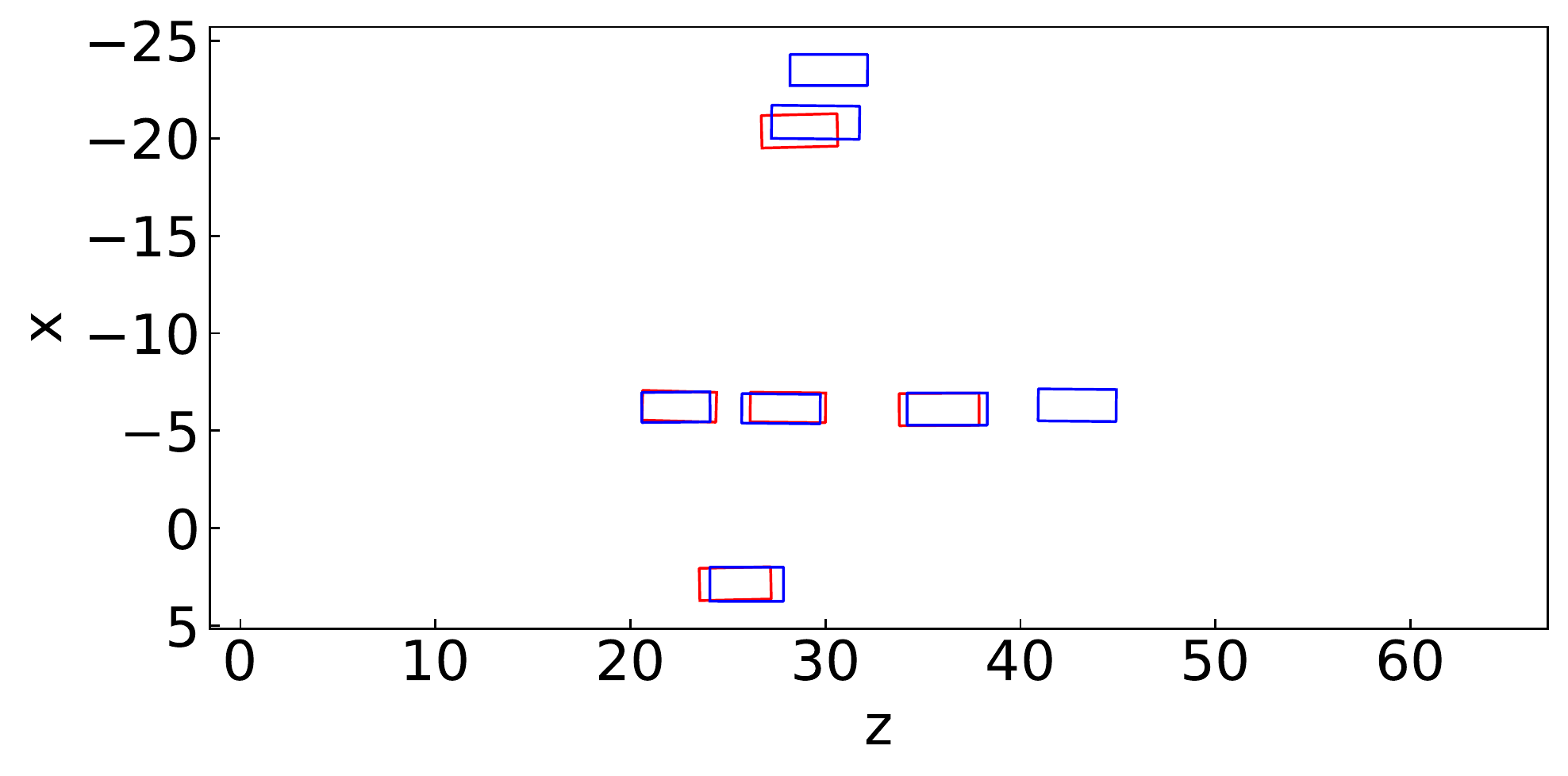}}

  \includegraphics[align=c,width=6.3cm]{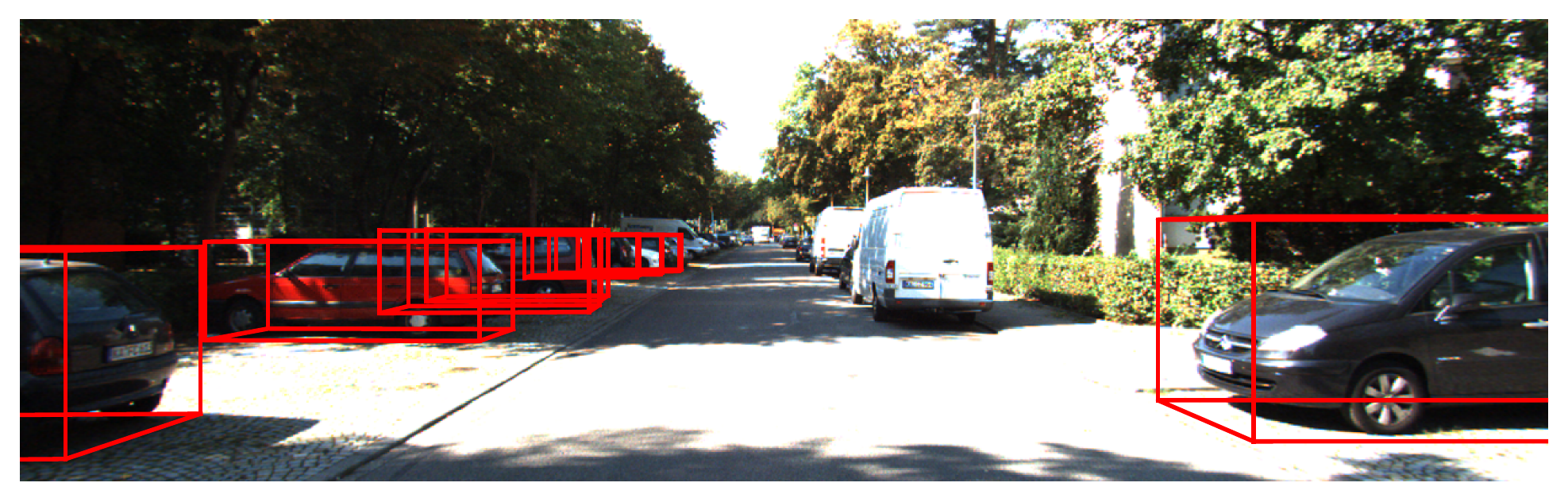}
  \includegraphics[align=c,width=6.3cm]{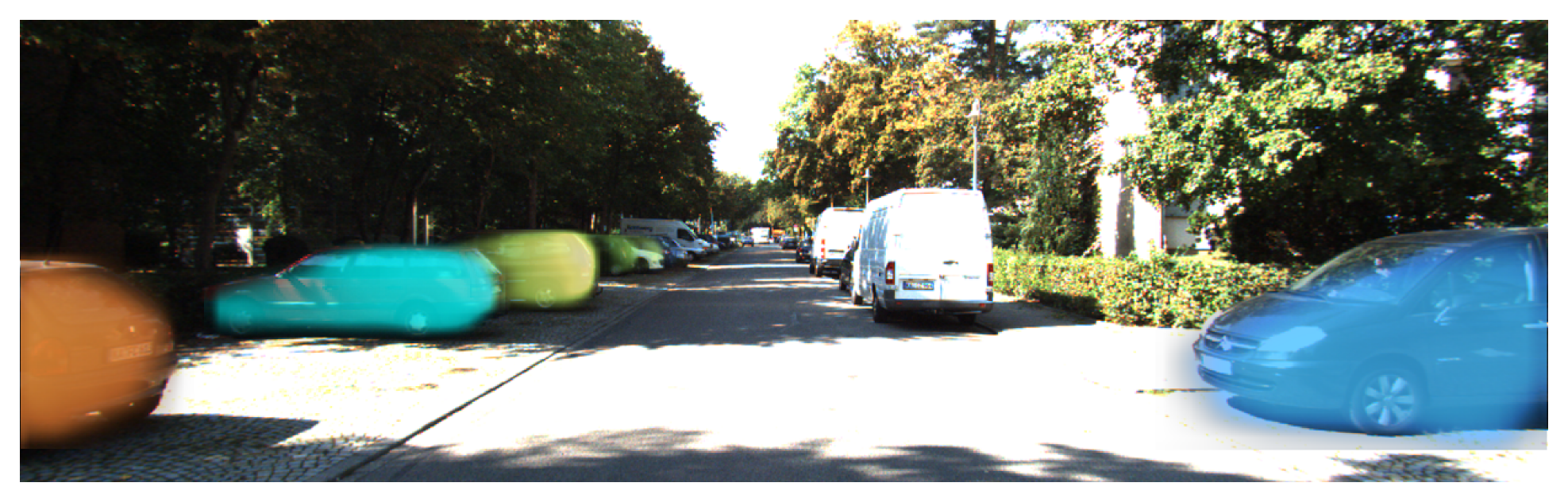}
  \raisebox{-0.18\height}{\includegraphics[align=c,width=5.0cm]{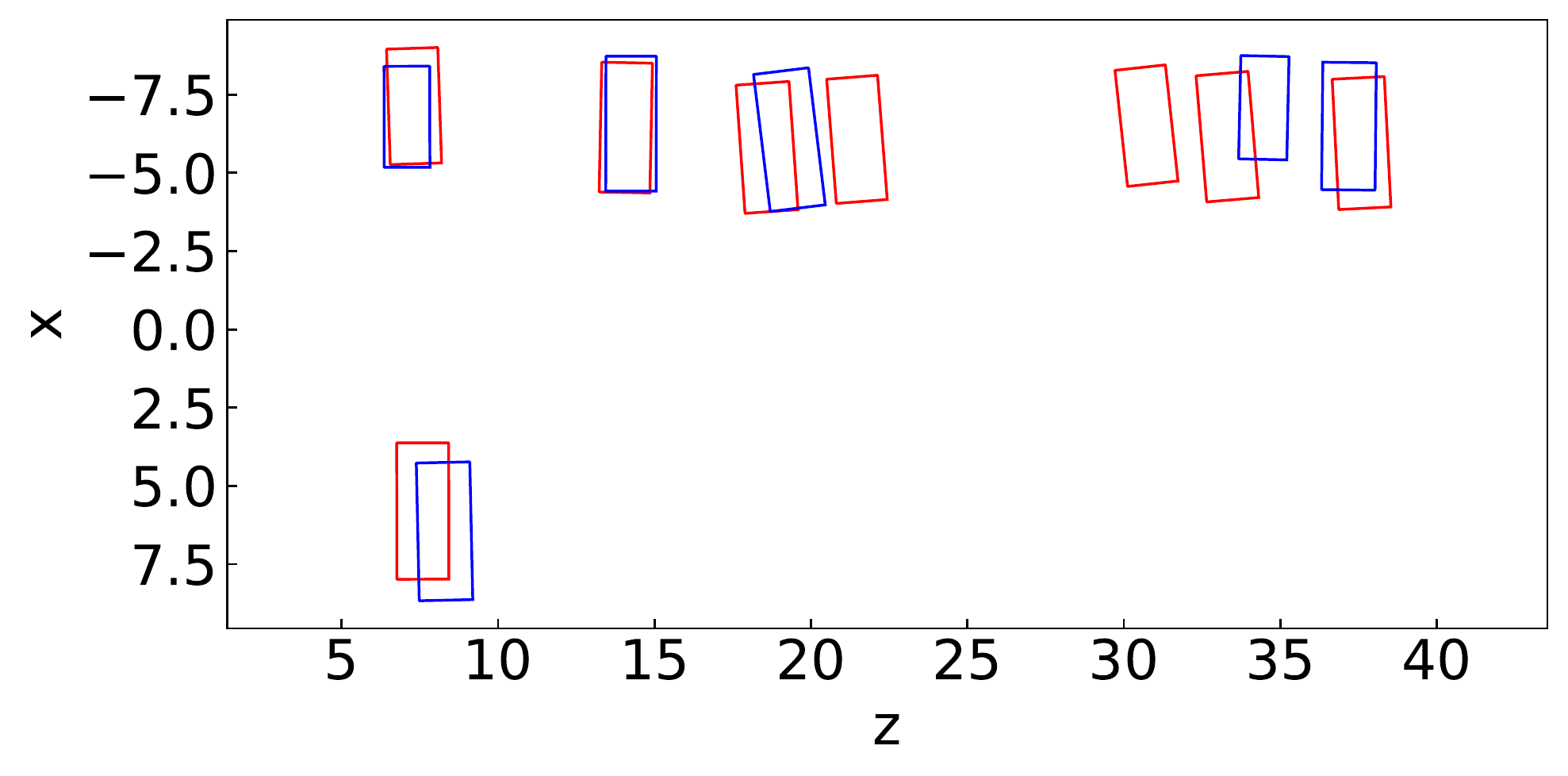}}

  \includegraphics[align=c,width=6.3cm]{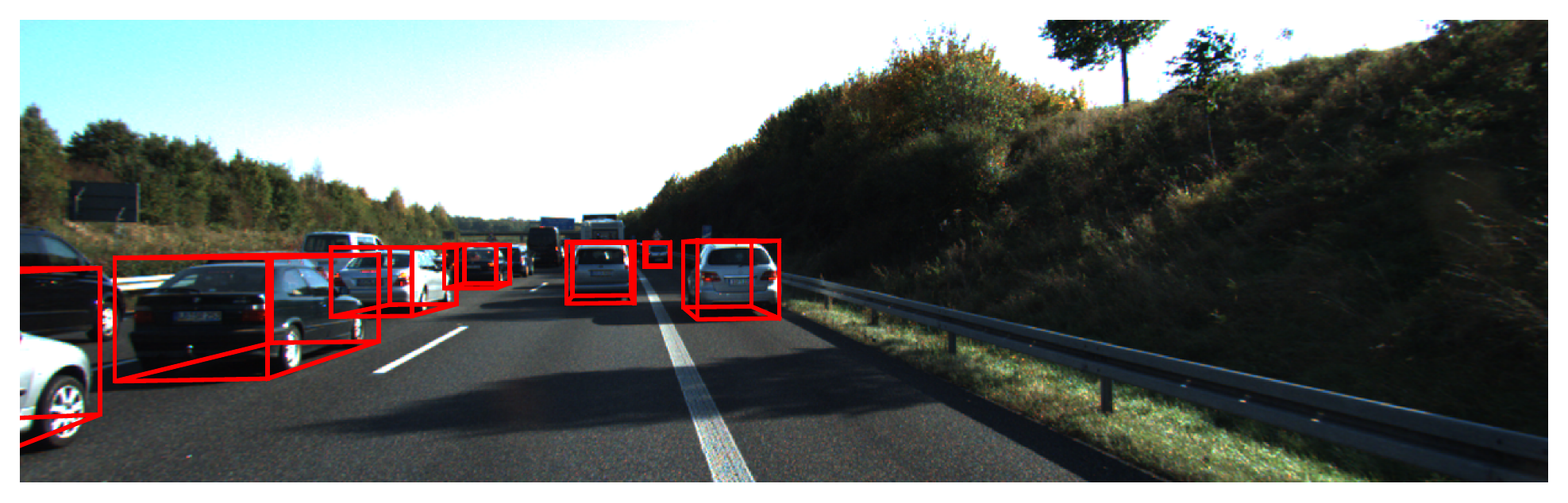}
  \includegraphics[align=c,width=6.3cm]{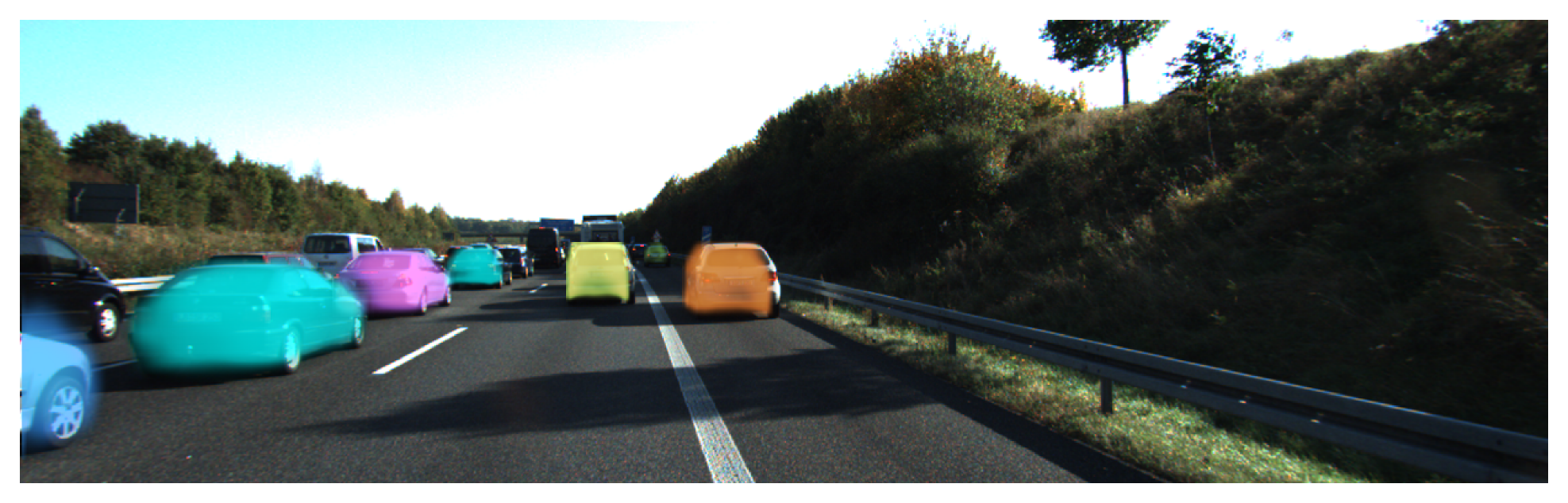}
  \raisebox{-0.18\height}{\includegraphics[align=c,width=5.0cm]{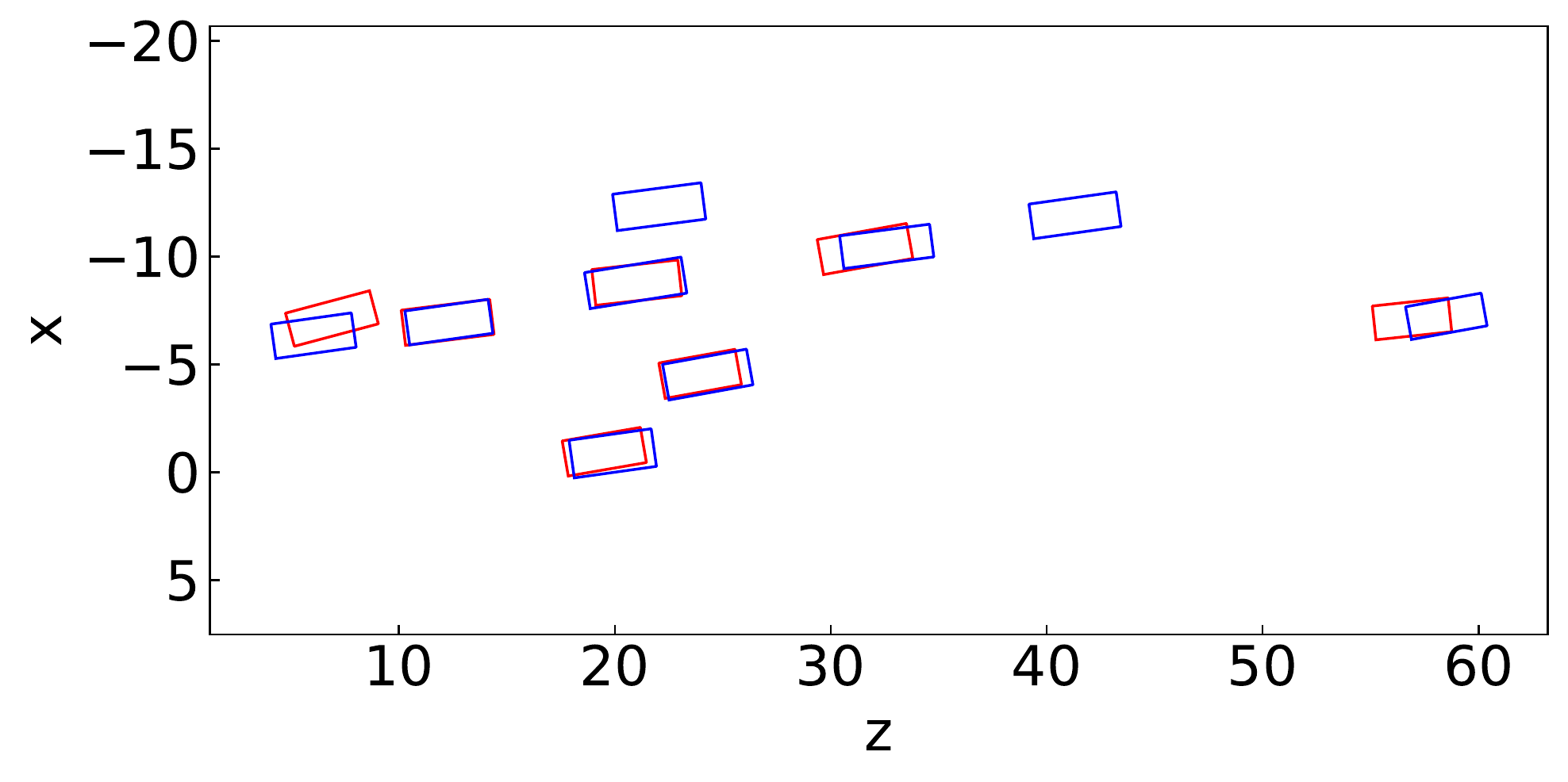}}

  \includegraphics[align=c,width=6.3cm]{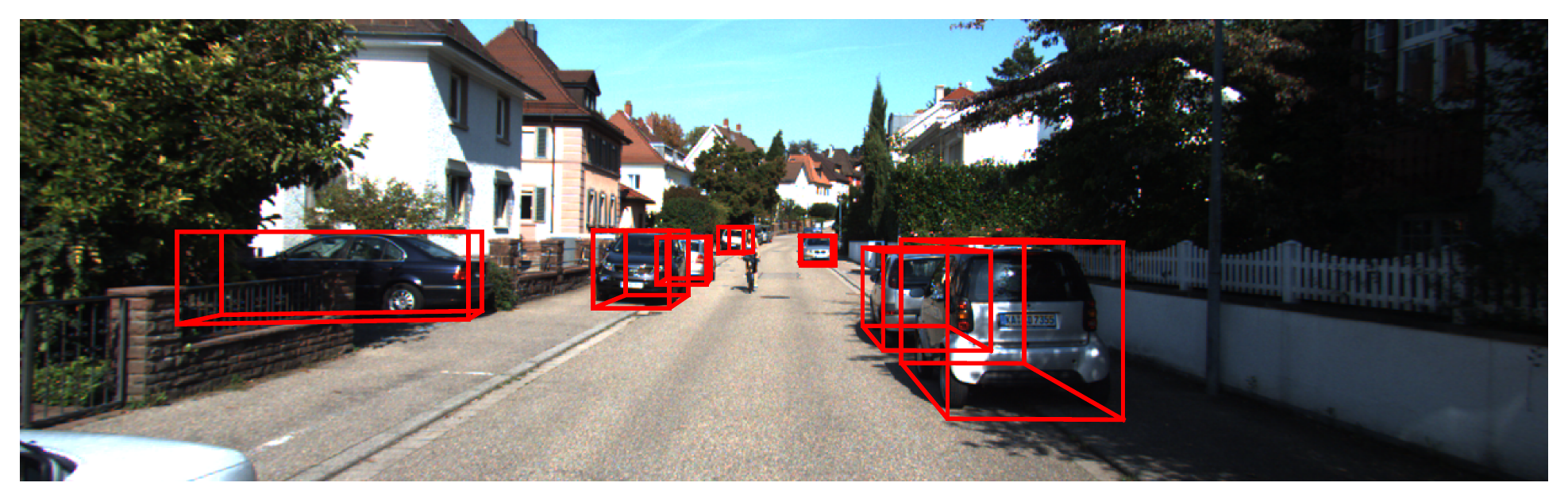}
  \includegraphics[align=c,width=6.3cm]{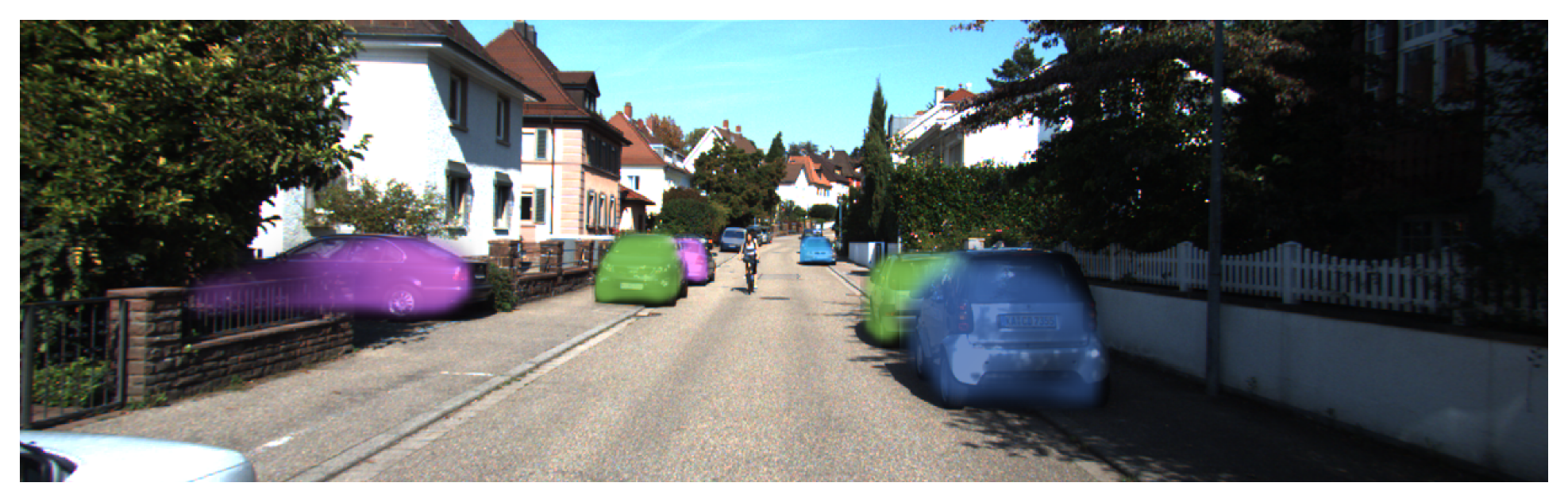}
  \raisebox{-0.18\height}{\includegraphics[align=c,width=5.0cm]{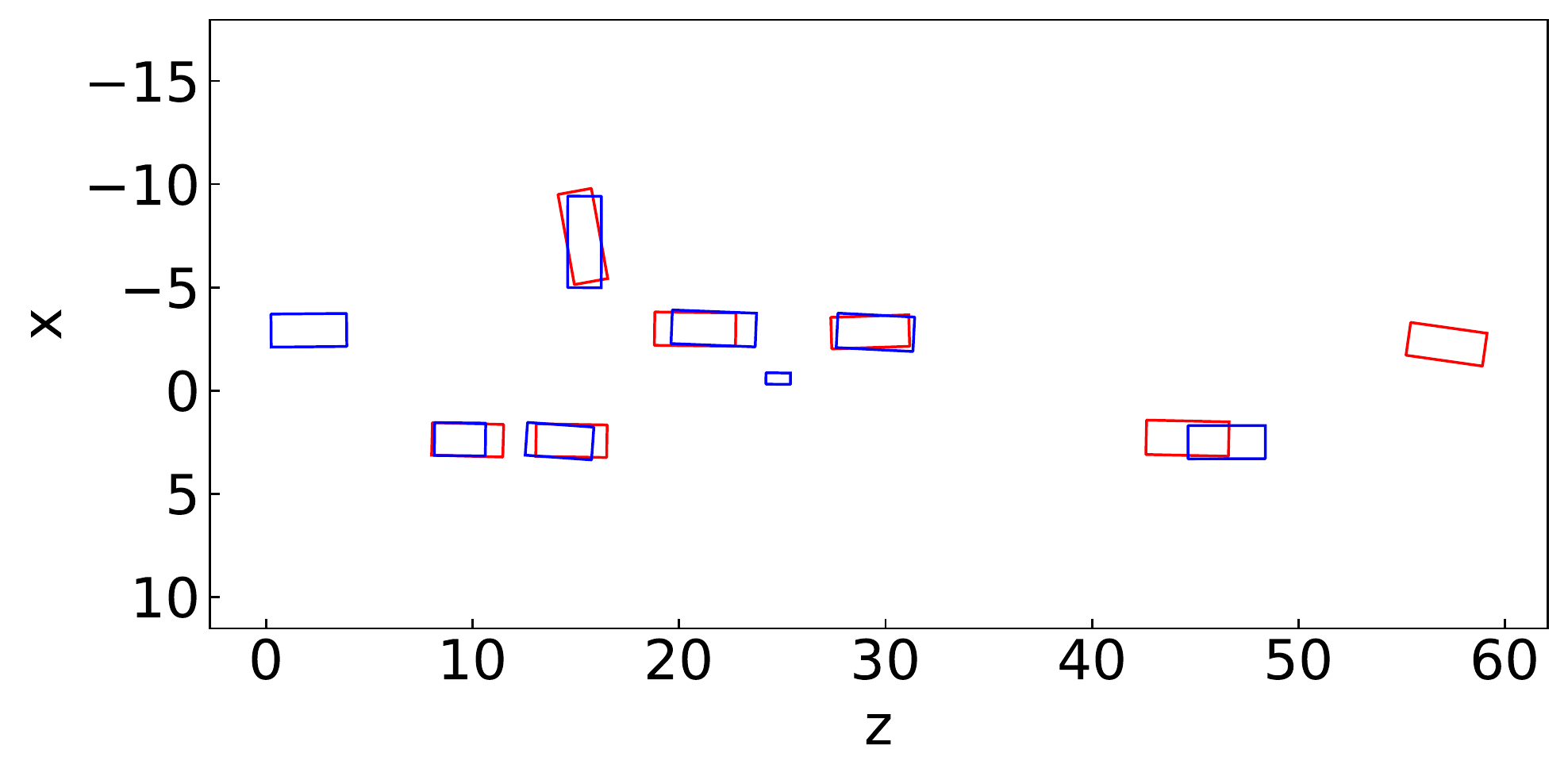}}
  
  \caption{Qualitative results on KITTI \textit{val} set. The \textit{1st} column shows the detected 3D boxes along with the objects. The \textit{2nd} column shows the visible area of objects predicted by the segmentation head. On the \textit{3rd} column, the detection results are shown along with the ground-truth in the BEV. The detected boxes and the ground-truth are in \textcolor{red}{red} and \textcolor{blue}{blue} respectively.\label{fig:qualitative_results}}
\end{figure*}

In this section, the performance of our shape-aware detector is studied on both the \textit{test} and \textit{val} set of 
 KITTI in comparison to state-of-the-art methods. The 3D average precision ${\text{AP}_{3D}}$ and the 
average depth similarity ${\text{ADS}}$ are reported for all the methods. Note that we only 
evaluate ${\text{ADS}}$ on the \textit{val} set since the ground-truth of 
the \textit{test} set is not available. Evaluation metrics are  
divided into easy, moderate, and hard settings according to the height, 
occlusion, and truncation level of objects. All evaluation metrics use 
\textit{40} recall points instead of \textit{11} recall points as recommended in 
\cite{simonelli2019disentangling}. 
As shown in Table~\ref{tab:performance}, our method in general outperforms the most recent method in the literature. In particular it outperforms MonoFlex on the \textit{test} set considerably. Moreover, our method shows consistently better accuracy in depth estimation than the rest of methods. Figure~\ref{fig:qualitative_results} shows the qualitative results on four monocular street views. The second column shows the results output from our segmentation head. The produced mask for a target object indicates the possibility of a pixel being on the object and the visibility of the object. For occluded objects, the masks highlight their visible areas, which enables the model to be free from the intereference of the context noises.

\subsection{Performance Analysis on Occluded Objects}
\begin{table}[htbp]
  \centering
  \caption{
    Performance on the occluded and fully visible objects. The evaluation is made on Cars of the KITTI \textit{val} set}
    \begin{tabular}{l|rr|rr}
    \toprule
    \multicolumn{1}{c|}{\multirow{2}[0]{*}{Approaches}} & \multicolumn{2}{c|}{${\text{AP}_{3D}}$}                           & \multicolumn{2}{c}{${\text{ADS}}$} \\
                                                        & \multicolumn{1}{c}{Fully Visible} & \multicolumn{1}{c|}{Occluded} & \multicolumn{1}{c}{Fully Visible} & \multicolumn{1}{c}{Occluded} \\
    \midrule
    MonoFlex* & 13.34          & 6.29          & 62.35          & 43.59 \\
    Ours      & \textbf{15.21} & \textbf{8.92} & \textbf{64.68} & \textbf{50.55} \\
    \bottomrule
    \end{tabular}%
  \label{tab:occlusion_analysis}%
\end{table}%

To further investigate the improvement that our method brings to the detection of occluded objects, we divide objects into two categories, the fully visible and the occluded, according to their occlusion state and truncation score. In KITTI, objects with 
occlusion state and truncation score equal to \textit{0} are marked as fully visible. The rest that their occlusion states are \textit{1} or \textit{2}, or truncation scores are larger than \textit{0} are marked as occluded. The objects whose occlusion state is \textit{3} (unknown) are ignored in the evaluation.

The evaluation results are reported in Table~\ref{tab:occlusion_analysis}. As seen from the table, the occluded objects are harder 
to be precisely located than the fully visible objects in the 3D space. Compared to MonoFlex, our method boosts the performance on 
the occluded objects by \textit{41.81\%} and \textit{15.97\%} in terms of the ${\text{AP}_{3D}}$ and the ${\text{ADS}}$  respectively.
Moreover, the shape-aware detector is also beneficial to the fully visible objects, for which \textit{14.02\%} and \textit{3.74\%} improvements are observed for the ${\text{AP}_{3D}}$ and the ${\text{ADS}}$ respectively.

\section{Conclusion}
\label{sec:conclusion}
We have presented a simple but effective auxiliary training task named shape-aware feature learning, which aims to improve the model performance on the occluded objects. With uncertainty weighted loss function, our method is able to learn from sparse and noisy segmentation labels, relieving of the laborious manual mask annotation. Considerable improvements are observed on the full-view objects as well as the occluded objects in particular. In addition, a metric called average depth similarity that is complementary to the current popular evaluation protocol is proposed to measure the performance of a monocular 3D detector. It allows a more comprehensive understanding about the monocular 3D object detection models.

\ifCLASSOPTIONcompsoc
 \section*{Acknowledgments}
\else
 \section*{Acknowledgment}
\fi
This work is fully supported by Ningbo Boden AI Technology Co., Ltd. from Ningbo, China. It is also supported by National Natural Science Foundation of China under grants 61572408 and 61972326.

\bibliographystyle{ieeetr}
\bibliography{refs}

\end{document}